\documentclass[letterpaper, 10 pt, conference]{ieeeconf}  

\IEEEoverridecommandlockouts                              

\usepackage{graphicx} 
\usepackage{float} 
\usepackage{subfig}
\usepackage{amsmath}
\usepackage{algorithm}
\usepackage{array}
\usepackage{algorithmic}
\usepackage{amsfonts}
\usepackage{amsmath}
\usepackage{amsthm,amsmath,amssymb}
\usepackage{mathrsfs}
\usepackage{cite}
\usepackage{balance}
\usepackage{bbding}
\usepackage{booktabs}
\usepackage{multirow}
\title{\LARGE \bf
Collective Behavior Clone with Visual Attention \\
via Neural Interaction Graph Prediction 
}

\author{Kai Li$^{1,2}$, Zhao Ma$^{2}$, Liang Li$^{3,4,5}$ and Shiyu Zhao$^{2}$
	\thanks{$^{1}$College of Computer Science and Technology at Zhejiang University, Hangzhou, China.
  }%
	\thanks{$^{2}$Department of Artificial Intelligence at Westlake University, Hangzhou, China.
		{\tt\small \{likai,mazhao,zhaoshiyu\}@westlake.edu.cn}}%
        \thanks{$^{3}$Department of Collective Behaviour, Max Planck Institute of Animal Behavior, Konstanz, Germany. {\tt\small lli@@ab.mpg.de}}
            \thanks{$^{4}$Centre for the Advanced Study of Collective Behaviour, University of Konstanz, Konstanz, Germany.}
            \thanks{$^{5}$Department of Biology, University of Konstanz, Konstanz, Germany.}
    	\thanks{$^{6}$https://github.com/WindyLab/CBC}%
}

\usepackage[mathscr]{eucal}
\begin{document}

	\maketitle
	\thispagestyle{empty}
	\pagestyle{empty}

	\begin{abstract}
In this paper, we propose a framework, collective behavioral cloning (CBC), to learn the underlying interaction mechanism and control policy of a swarm system. Given the trajectory data of a swarm system, we propose a graph variational autoencoder (GVAE) to learn the local interaction graph. Based on the interaction graph and swarm trajectory, we use behavioral cloning to learn the control policy of the swarm system. To demonstrate the practicality of CBC, we deploy it on a real-world decentralized vision-based robot swarm system. A visual attention network is trained based on the learned interaction graph for online neighbor selection. Experimental results show that our method outperforms previous approaches in predicting both the interaction graph and swarm actions with higher accuracy. This work offers a promising approach for understanding interaction mechanisms and swarm dynamics in future swarm robotics research. Code and data are available$^{6}$.
	\end{abstract}

	\section{Introduction}
Behavioral cloning (BC) is a supervised learning approach that enables agents to replicate expert behaviors. In this paper, we extend BC to address the problem of collective behavioral cloning (CBC), which focuses on learning the control policy of a swarm system based on its observed trajectories. The goal of CBC is to identify and replicate the underlying mechanisms that govern the formation of collective behaviors in swarms. This approach has significant implications for understanding and predicting complex group dynamics, such as crowd movement in urban environments. Moreover, CBC provides valuable insights into natural swarm systems, including animal groups like bird flocks or fish schools, offering a novel perspective on the emergence of collective intelligence.

   \begin{figure}[thpb]
      \centering
      \includegraphics[scale=0.8]{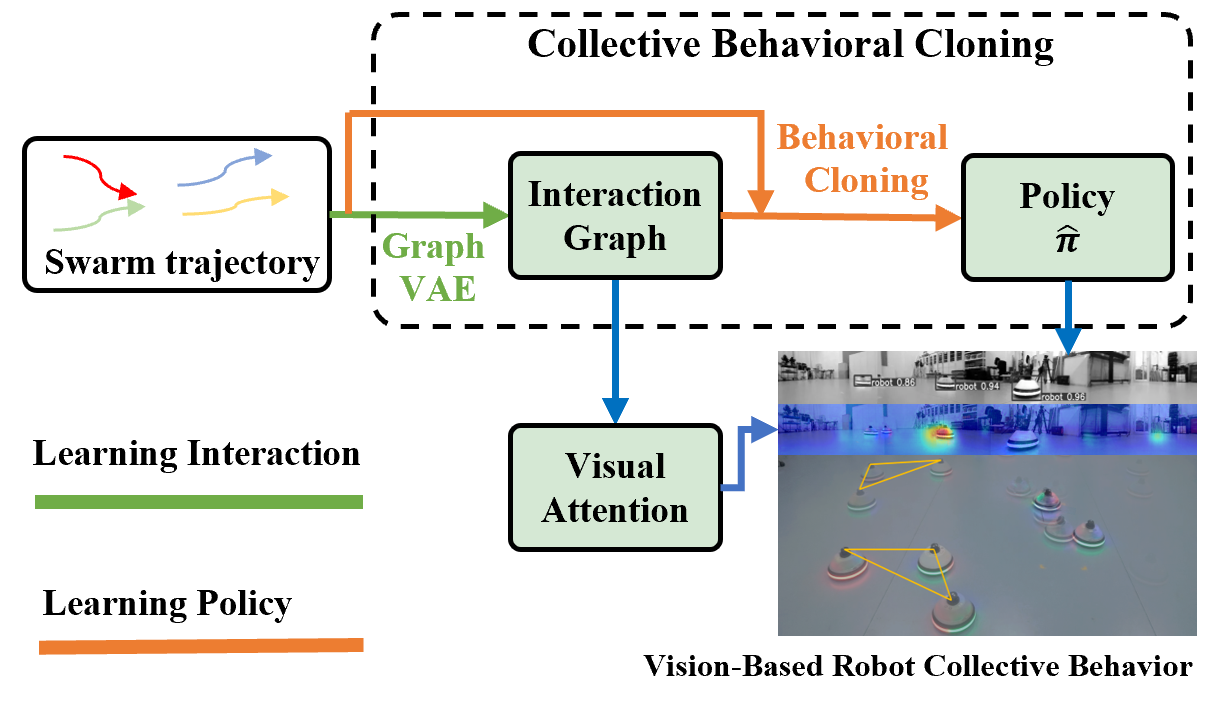}
      \caption{Overview of the CBC framework. The GVAE learns the interaction graph from the trajectory and behavior cloning learns the control policy. Then collective behavior of the swarm can be replicated. Since we use a vision-based decentralized robot swarm system with no wireless communication, a visual attention network is trained based on the learned interaction graph for online interaction robots selection. }
      \label{figurelabel}
   \end{figure}

In contrast to BC, CBC presents greater learning challenges due to the unknown local interaction mechanisms in swarm systems. Knowing which neighboring agents are interacting with each other is crucial for the collective behavior policy.
BC focuses solely on learning the expert's policy, whereas CBC must learn both the expert's policy and the local interaction mechanism at a given time.
The key challenge of CBC lies in the fact that the local interaction mechanisms of a swarm system cannot be directly observed. Several local interaction mechanisms have been proposed in previous research, such as range-based models, topological models, and visual attention\cite{zheng2024emergence,li2024collective,schilling2022scalability}, which aim to achieve swarm behaviors. However, these models are typically human-designed and may not be consistent with the actual interaction mechanisms within a natural swarm.

To address the challenge of understanding local interaction mechanisms and learning the underlying policy behind the collective behavior, we propose the CBC framework in this paper. We propose a graph variational autoencoder (GVAE) model to learn the local interaction mechanism. Based on the learned interaction mechanism, we select interacting neighbors and learn the control policy that governs the swarm’s collective behavior.
We also design a vision-based decentralized robot swarm system and demonstrate CBC in practice. In this system, the visual attention network is used for online selection of interacting robots, as vision is the only sensory modality available in our decentralized setup.

The three main contributions of this paper are as follows:

1) We propose a novel framework, collective behavioral cloning (CBC), to learn both the interaction mechanisms and control policy from swarm trajectory data. Our method shows higher accuracy in terms of action and trajectory error thanks to the learned interaction graph.

2) We enhance the GVAE model for improved interaction graph prediction. Our method outperforms state-of-the-art methods, demonstrating higher precision in predicting interaction graphs within swarm environments.

3) We deploy the proposed framework on a real-world vision-based robotic swarm system, effectively demonstrating its feasibility and practicality in challenging scenarios that involve wireless communication denial and full decentralization.

\section{Related Works}
\textbf{Modeling Swarm Behaviors}.
In swarm robotics, researchers have extensively explored collective behaviors from the perspectives of physics and biology to understand the mechanisms behind coordinated movement. This behavior typically unfolds in two stages: the neighbor selection stage and the motion decision stage \cite{zheng2024emergence, li2024collective}. Numerous classical models have been proposed to explain motion decision-making. For instance, the model in \cite{vicsek1995novel} uses the average direction of neighbors to form a self-propelled system, while \cite{couzin2002collective} introduces interaction rules based on separation, alignment, and cohesion (SAC) to model swarm dynamics.
The neighbor selection stage involves choosing which agents to interact with. Various human-designed models have been proposed to model this process, such as range-based, K-nearest, and vision-based selection rules \cite{zheng2024emergence, li2024collective, schilling2022scalability}. These interaction selection models and control policies have been applied in practical robot swarms \cite{huang2024collision, soria2022distributed, batra2022decentralized, schilling2021vision, schilling2019learning}.
In this paper, we differ from the traditional first-principle-based modeling by employing learning-based methods to discover the underlying interaction patterns and motion decision processes directly from trajectory data of swarm systems. This approach allows us to replicate and deepen our understanding of collective behaviors.

\textbf{Neural Interaction Graph Prediction}.
Modeling the hidden interaction in multi-agent systems with graph representation learning has made tremendous progress in recent studies\cite{liu2023intention,huang2021learning,vemula2018social,yu2020spatio,sun2022interaction,kipf2018neural,graber2020dynamic,dax2023disentangled}. Graph models can learn inter-agent relations from swarm trajectories. In this way, the underlying interaction pattern of the swarm system can be revealed.
Neural interaction graph prediction can be mainly divided into two groups. One group\cite{kipf2018neural, graber2020dynamic,sun2022interaction,dax2023disentangled} combines graph neural network with the variational autoencoder (VAE) to infer the relations and predict trajectories in a multi-agent system. The other group\cite{ liu2023intention,huang2021learning,vemula2018social,yu2020spatio} uses transformer architecture and multi-head attention to predict agent trajectories while modeling interactions in multi-agent systems. 
In this work, we propose a model based on the graph VAE and achieve better accuracy for graph prediction in the swarm setting.

\textbf{Visual-Attention for Robotics}.
Visual attention is a fundamental feature of visual sensing that actively filters out the task-relevant area within an image. Deep neural networks\cite{islam2022svam} have been proposed to imitate the visual attention mechanism. Visual attention has been applied to various robotic tasks as well. Some works use visual attention as an enhancement for trajectory planning \cite{wapnick2021trajectory}, cooperative drone control\cite{yin2023towards}, and autonomous driving\cite{liu2020using}. 
Incorporating visual attention into decentralized vision-based swarms has not yet been fully explored in the literature. In this work, since the vision-based robot swarm system is decentralized and wireless communication is not available, we use the visual-attention network to select interaction targets during the online running.
   
\section{Problem Formulation}
A multi-agent system can be viewed as a spatio-temporal graph $G$, where each agent represents a node, and edges are formed between agents that interact with each other. 
In swarm robotics, the control policy that drives collective behavior is typically formulated as:
\begin{align}
\pi(\mathbf{s}) = f\left( \left\{ \mathbf{s}_j \mid j \in I_{i} \right\} \right)
\end{align}
where $\mathbf{s}$ is the states of the agents that belong to the interaction agents set $I_{i}$ of the $i$th agent. Here, $I_{i}$ refers to the set of agents that are connected to the $i$th agent in the graph $G$. In the following part of this paper, we refer to $I_{i}$ as the interaction graph, which represents the local connectivity structure of the 
$i$th agent within the spatio-temporal graph $G$.
In the context of swarm robotics\cite{zheng2024emergence,li2024collective,fine2013unifying}, swarm control policy is also divided into two stages, namely the neighbor selection stage and the motion decision stage. The  
neighbor selection stage refers to obtaining $I_{i}$ with a certain neighbor selection rule. The motion decision stage refers to generating the action with $\pi(\mathbf{s})$ based on $I_{i}$.
Here the states $\mathbf{s}$ can be the relative position or velocity between the agents. The policy $\pi$ outputs an action that governs the collective behavior of the swarm, such as flocking, milling, or swarming.
To summarize, the goal of CBC is to learn the policy 
$\pi$ and interacting graph $I_{i}$ from a given trajectory data of a swarm system. 

\section{Method}
\subsection{Overview} 
The CBC framework consists of two primary phases: 1) learning the local interaction graph and 2) learning the control policy.
 
     \begin{figure}[thpb]
      \centering
      \includegraphics[scale=0.9]{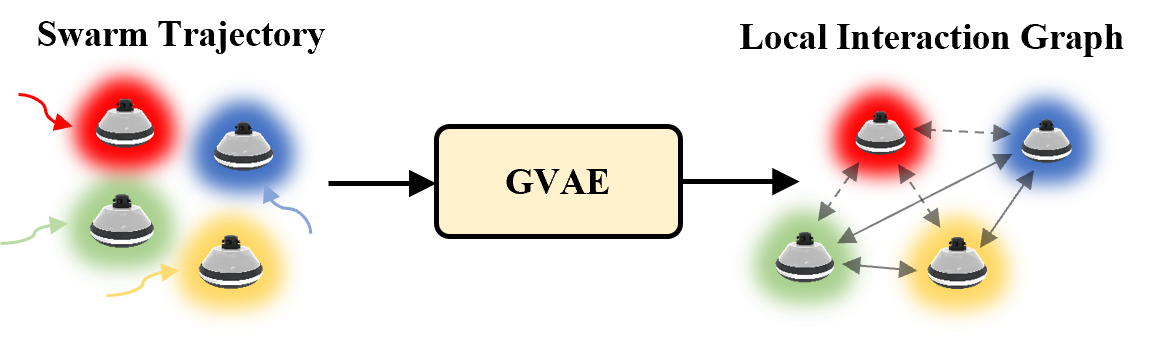}
      \caption{The GVAE model takes the swarm trajectory and outputs the local interaction graph edges.}
      \label{figureLearg Graph}
   \end{figure}

\textbf{Learning the Interaction Graph}. A multi-agent system can be interpreted as a dynamic graph in space and time, where each agent is represented as a node, and interactions between agents are represented by edges connecting them. Given the trajectories of the swarm system, our goal is to learn which agents are connected and interacting at any given time. We propose a Graph Variational Autoencoder (GVAE) model that utilizes a graph neural network to extract features from historical trajectories. The VAE then predicts the edges of the graph, representing the interactions between agents. This process also corresponds to the neighbor selection stage in swarm robotics\cite{zheng2024emergence,fine2013unifying}.
These predicted edges determine which neighboring agent's states should be fed into the policy for decision-making. The detailed structure of the GVAE model is introduced in Section IV-B.

\textbf{Learning the Control Policy}.
The control policy learning module follows the general behavioral cloning framework. Since a multilayer perceptron (MLP) can serve as a universal function approximator, we use an MLP as the action network for behavioral cloning: 
\begin{align} \hat{\pi}(\theta;\mathbf{s}) = \text{MLP}\left( \{ \mathbf{s}_{j} \mid j \in I_{i}  \} \right), 
\end{align} 
where $\hat{\pi}(\theta;\mathbf{s})$ is the learned control policy, and $\theta$ represents the MLP parameters. $\hat{\pi}(\theta;\mathbf{s})$ takes the states of the selected neighbors in $I_{i}$ and outputs the action for the agent. 
The loss function for control policy learning is the mean squared error between the velocity action predicted by the MLP and the actual velocity action in the swarm trajectory data, governed by $\pi(\mathbf{s})$. 
\begin{align}
\mathcal{L} = ||\hat{\pi}(\theta;\mathbf{s}) -  \pi(\mathbf{s})||_{2}.
\end{align}
This process also corresponds to the motion decision stage in swarm robotics.

\subsection{Neural Interaction Graph Prediction}
     \begin{figure}[thpb]
      \centering
      \includegraphics[scale=0.72]{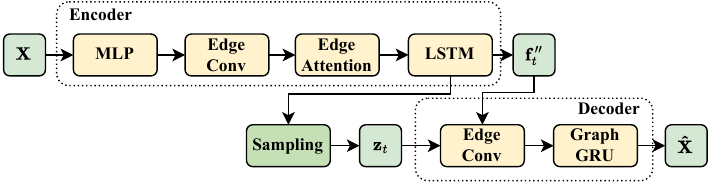}
      \caption{Structure of the GVAE model. $\mathbf{X}$ and $\mathbf{\hat{X}}$ denotes the input and predicted trajectory respectively. $\mathbf{z}$ represents the interaction graph edge logits.}
      \label{figureGAVE}
   \end{figure}
\textbf{Overview}. We use a GVAE model to learn the local interaction graph. Structure of the GVAE is shown in Fig.~\ref{figureGAVE}. The goal of GVAE is to learn the underlying interaction graph edges from the global trajectory of the swarm. The trajectory of the swarm is defined as 
$\mathbf{X} \in \mathbb{R}^{\mathrm{T \times N \times F}}$,
where $\mathrm{T}$ denotes the sequence length, $\mathrm{N}$ denotes the number of agents, and $\mathrm{F}$ denotes the feature dimension. The trajectory feature of each agent contains its position and velocity in the global frame. The latent interaction graph is defined as $\mathbf{Z} $, 
which is represented by a probability adjacent matrix at each timestamp. 
The key novel features of our model are as follows.
First, we apply an attention mechanism across node-level embedding to model the relative importance. Additionally, we leverage the modern graph edge convolution operators for feature extraction in both the encoder and decoder to achieve better expressiveness.

\textbf{Encoder}. The GVAE employs an encoder-decoder structure. First, a multi-layer perceptron (MLP) extracts the node embedding $\mathbf{f}_{t}$ from the raw swarm trajectory. This node embedding is then encoded using an edge convolution layer \cite{wang2019dynamic},
\begin{align}
\mathbf{f}_{i,t}^{\prime}=\sum_{i \in \mathcal{N}_{t}(i)} h_{\theta}(\mathbf{f}_{i,t} || \mathbf{f}_{j,t}-\mathbf{f}_{i,t})
\label{projection_equation},
\end{align}
where $i$ denotes the $i$th agent, $h_{\theta}$ is a MLP network, $||$ denotes the concatenation of the two embedding, and $\mathcal{N}_{t}(i)$ denotes the graph neighbors. We set the graph to be fully connected at this stage.

Since the interaction graph operates at the edge level, we convert the node-level embedding into an edge-level embedding and apply a dot-product attention mechanism to model potential interactions between the $i$th and $j$th agents,
\begin{align} \mathbf{e}_{ij,t}^{\prime} = \text{Attention}(\mathbf{f}_{i,t}^{\prime} || \mathbf{f}_{j,t}^{\prime}). \label{1st} \end{align}

To handle time-varying interaction graphs, the node-level embedding $\mathbf{f}_{t}^{\prime}$ obtained from the graph network block is fed into a long short-term memory (LSTM) module to further extract time-specific features $\mathbf{f}_{t}^{\prime\prime}$ and latent graph logits $\mathbf{e}_{t}^{\prime\prime}$,
\begin{align} \mathbf{f}_{t}^{\prime\prime}, \ \mathbf{e}_{t}^{\prime\prime} = \text{LSTM}(\mathbf{f}_{t}^{\prime}) \label{projection_equation}. 
\end{align}

\textbf{Sampling}. The encoder outputs latent graph logits for potential interactions. To enable gradient-friendly categorical sampling, we follow the standard VAE approach and use Gumbel-Softmax \cite{jang2016categorical} to obtain the sampled latent edge probabilities between the $i$th and $j$th agents,
\begin{align}
\mathbf{z}_{ij,t}= \rm{SOFTMAX}((\mathbf{e}_{ij,t}^{\prime\prime} + \mathbf{u})/\tau),
\end{align} \label{2nd}
where $\mathbf{u}$ is a sample vector from Gumbel distribution\cite{jang2016categorical} and temperature $\tau$ controls the smoothness.

\textbf{Decoder}. The decoder takes the sampled latent interaction graphs $\mathbf{z_{t}}$ and the node embedding $\mathbf{f_{t}^{\prime\prime}}$. These features are fed into a weighted edge convolution layer,
\begin{align}
\mathbf{f}_{i,t}^{\prime\prime\prime}= \sum_{i \in \mathcal{N}_{t}(i) | \mathbf{z}_{t}}h^{\prime}_{\theta}(\mathbf{f}_{i,t}^{\prime\prime} || \mathbf{f}_{j,t}^{\prime\prime})\label{edge_decoder},
\end{align}
to extract the decoded node-level embedding, where we use the edge probabilities $\mathbf{z}_{t}$ from the encoder as weights for aggregation. The node-level embedding is then fed into a graph gated recurrent unit (GRU) to recover the trajectories.

\textbf{Loss Functions}. The loss function follows the general form of VAE and consists of a reconstruction term and a prior term.
The reconstruction term measures the difference between the reconstructed trajectory and the ground truth,
\begin{align}
\mathcal{L}_{re} = \frac{1}{\rm{T}}\sum_{t \in \rm{T} }\sum_{i}\frac{1}{2\sigma^{2}}(\mathbf{x}_{t,i}-\mathbf{\hat{x}}_{t,i})^2\label{NLL_loss},
\end{align}
where $\sigma^2$ is a hyperparameter that reflects the prior variance of agent trajectories. The prior term measured by KL-divergence
regulates the distribution of the latent variables:
\begin{align}
\mathcal{L}_{pr} = D_{KL}(p(\mathbf{z}_{t}|\mathbf{X}) || p(\mathbf{z}_{t}|\mathbf{X},\mathbf{z}_{1:t-1})).
\end{align}

\subsection{Vision-Based Robot Swarm System}
     \begin{figure}[thpb]
      \centering
      \includegraphics[scale=0.7]{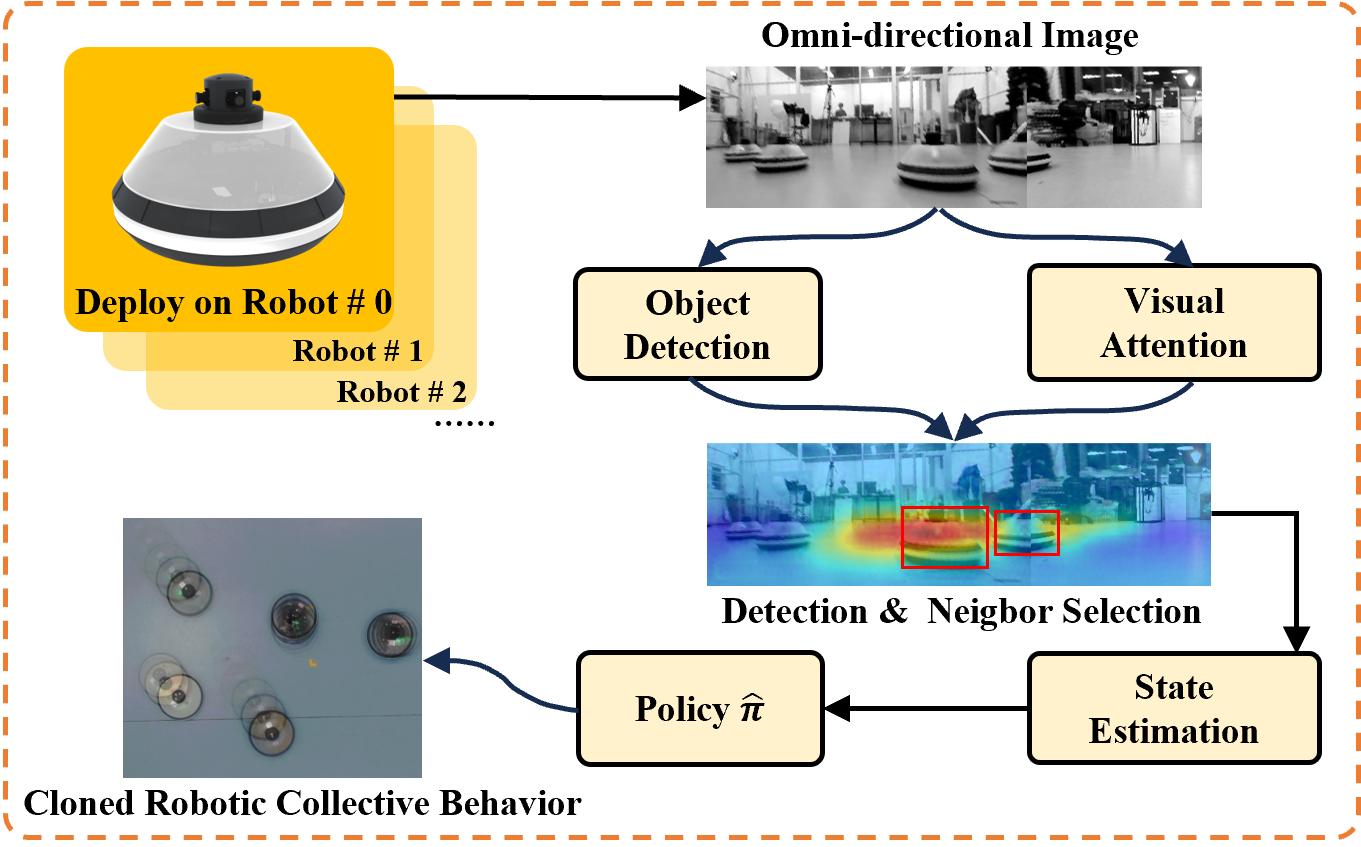}
      \caption{Architecture of the vision-based robot swarm. The visual attention network is trained using labels generated from the graph edge predictions by the GVAE. Since the system operates in a decentralized manner and wireless communication is not allowed, visual attention is leveraged to select robots for engagement.}
      \label{figureRobotSwarmSystem}
   \end{figure}
   
\textbf{Overview} To test the proposed CBC framework in practice, we design a vision-based robot swarm system. The system structure is shown in Fig~\ref{figureRobotSwarmSystem}.
In practice, since the robot swarm operates in a decentralized manner with no inter-robot communication, the selection of interaction neighbors must be achieved through vision. Neighbor selection in the vision-based robot swarm is handled by a visual attention network and an object detection network. Specifically, we train a U-Net \cite{ronneberger2015u} as the visual attention network to predict regions of the image that receive the highest attention, and a YOLOv5 \cite{redmon2016you} network to detect other robots. The YOLOv5 model detects all robots within the image, while the visual attention network predicts areas of high interest. Robots detected within these high-attention areas are then selected as the target robots to engage with.
In summary, the visual attention network acts as a proxy for the interaction graph learned by the GVAE during online operation, enabling neighbor selection without relying on communication.

\begin{figure}[thpb]
      \centering
      \includegraphics[scale=1]{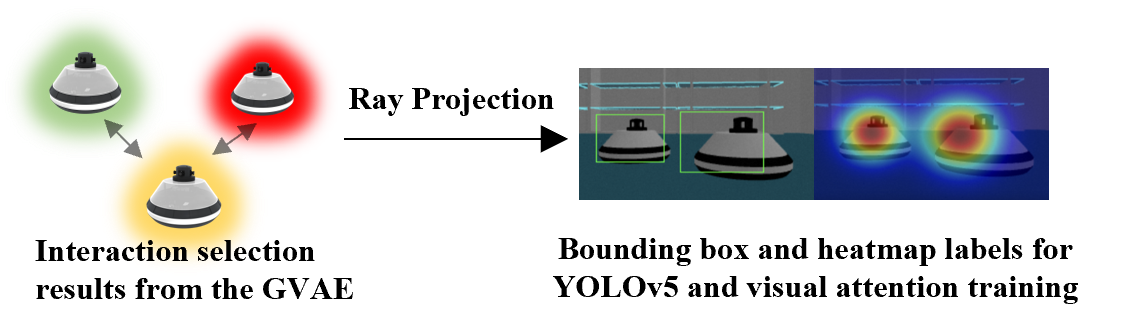}
      \caption{The GVAE predicts the interaction graph. For robots that have interactions, we project their 3D position onto the image space to create training labels for YOLOv5 and visual attention network.}
      \label{figureProjection}
   \end{figure}

To ensure that the neighbor selection is consistent with the learned interaction graph from the GVAE, the training labels for the visual attention network are generated using both the interaction prediction results and the robot trajectory. Given some swarm trajectory data, which represents the collective behavior to be cloned, the following process is used for visual attention network training:
1) The swarm is run according to the trajectory, with image and trajectory data being recorded and synchronized.
2) The GVAE is applied to obtain the interaction graph and neighbor selection results at each timestamp.
3) For each pair of robots connected in the interaction graph, based on their relative poses and camera parameters, each robot is projected onto the other's image space. 
4) A Gaussian blur is then applied at the projection point to create a heatmap label.
5) These heatmap labels are used to train the visual attention network. Kullback–Leibler divergence between the heatmap labels and predicted attention map is used as the training loss. The kernel size $s$ of the Gaussian blur for heat map is $s=\lambda/d$, where $d$ is the relative distance and $\lambda$ is a scaling factor. This indicates that the closer the interacting robot is, the larger the kernel size will be in the heatmap. Fig.~\ref{figureProjection} shows the projection process.

Since the swarm control policy relies on the states of interacting robots, i.e. the relative positions and velocities, a Kalman state estimator is used to estimate the states of high-attention interacting robots. Using the estimated motions, the learned policy $\hat{\pi}$ outputs actions that generate the collective behavior to be cloned.

\textbf{State Estimation}. Since the swarm control policy relies on the states of interacting robots, we introduce how to estimate the robot states from image bounding box observations here. 
The state estimation follows the pseudo-linear Kalman filter (PLKF) based on visual bounding box observations\cite{Ning2024}. The target state $\mathbf{x}=[\mathbf{p}^{\rm{T}},\mathbf{v}^{\rm{T}}]^{\rm{T}} \in \mathbb{R}^{6}$, where $\mathbf{p}$ is the relative position and $\mathbf{v}$ is the relative velocity. 
The state transition and process covariance are given by
\begin{align}
\mathbf{x}_{t+1}=\mathbf{F}\mathbf{x}_{t}, \ \mathbf{F}= \begin{bmatrix} \mathbf{I}^{3\times3} & dt\mathbf{I}^{3\times3} \\ \mathbf{0}^{3\times3} & \mathbf{I}^{3\times3} \end{bmatrix},
\\
\mathbf{Q}=diag(\sigma_{p}^{2},\sigma_{p}^{2},\sigma_{p}^{2},\sigma_{v}^{2},\sigma_{v}^{2},\sigma_{v}^{2}) \in \mathbb{R}^{6\times6},
\end{align}
where $\sigma_{p}^{2}$ is the position variance, $\sigma_{v}^{2}$ is the velocity variance.
The observations, i.e. the bearing vector $\mathbf{g}$ and target field angle $\theta$ with noise can be extracted from the bounding box and camera parameters:
\begin{align}
\mathbf{g} = \frac{\mathbf{p}} {  \lVert \mathbf{p} \lVert } + \boldsymbol{\mu}, \
      \theta=\frac{\ell} { \lVert \mathbf{p} \lVert }  \ + \ \omega ,\label{ob}
\end{align}
where $\ell$ is the physical diameter of the target, $\boldsymbol{\mu}$ and $\omega$ are noises in zero-mean Gaussian distribution with the covariance of $\Sigma_{\boldsymbol{\mu}}=\sigma_{\mu}^2\mathbf{I} \in \mathbb{R}^{3\times3}$ and $\sigma_{\omega}^2 \in \mathbb{R}$.
We choose the target field angle $\theta$ to represent the relative size measurement because $\theta$ remains invariant to camera rotation\cite{Ning2024}.
The observation equation of PLKF is 
\begin{align}
\mathbf{z} = \mathbf{H}\mathbf{x} + \boldsymbol{\nu}, \label{ob_eq} \
\mathbf{H}=[\mathbf{I}^{3\times3} \ \mathbf{0}^{3\times3}]\in \mathbb{R}^{3\times6},
\end{align}
where $\mathbf{z}=(l/\theta)\mathbf{g}\in\mathbb{R}^{3}$. The observation noise $\boldsymbol{\nu}$ is computed by combining Eq.~\ref{ob} and Eq.~\ref{ob_eq}: 
\begin{align}
\boldsymbol{\nu}=\lVert \mathbf{p} \lVert(\boldsymbol{\mu} - \frac{\omega \mathbf{g}}{\theta})\in \mathbb{R}^{3}.
\end{align}
Since $\boldsymbol{\nu}$ can be viewed as a linear combination of Gaussian noise $\boldsymbol{\mu}$ and $\omega$, the observation covariance is given by
\begin{align}
    \mathbf{R}=  
    \lVert \mathbf{p} \lVert^{2}(\Sigma_{\boldsymbol{\mu}} + \frac{\sigma_{\omega}^{2}}{\theta^{2}}\mathbf{g}\mathbf{g}^{\rm{T}}) \in \mathbb{R}^{3\times3}.
\end{align}

\section{Experimental Evaluations}
\subsection{Results of Neural Interaction Graph Prediction}
In this part, we aim to evaluate the interaction graph prediction ability of the GVAE model.

\textbf{Data}. We simulate the trajectory of a robot swarm with pre-defined interaction graph and control policy to build a custom benchmark dataset. In the simulation, a leader moves in a pre-defined pattern such as a circular, a linear motion, or a U-turn, and other agents follow its movements to form flocking behavior. A potential-field-like policy is employed to create flocking behavior for followers,
\begin{align}
\pi(\mathbf{r}_{ij}) =  \sum_{j \in I_{i}}(1-\frac{k^{3}}{\lVert \mathbf{r}_{ij} \lVert^{3}})\mathbf{r}_{ij},\label{flocking}
\end{align}
where $\mathbf{r}_{ij}$ denotes the relative position between the $i$th and $j$th agent. $k$ is a scaling factor, and $I_{i}$ is the interacting graph. 
The interaction graph across agents is evenly generated with range-based\cite{fine2013unifying,huang2024collision}, K-nearest\cite{fine2013unifying,soria2022distributed,huang2024collision}, and vision-only methods\cite{schilling2021vision,schilling2022scalability}. In this way, we have the ground truth result of which agents are interacting with each other at a given time. By comparing the predicted interaction graph edge with the ground truth data, we can evaluate the performance of our model.
 The initial positions of all agents are randomly generated in a 6m$\times$6m area. The control policy drives each agent to its migration destination while avoiding collision, forming a flocking behavior. We set the maximum speed of each agent at 0.2 m/s. The number of agents ranges from 3 to 9. 50K of simulations are generated for each experiment. The training, validation, and testing data are with a ratio of 80/10/10.

\textbf{Implementation Details}. We train our GVAE model for 100 epochs on an RTX 4090 GPU, using the Adam optimizer with a learning rate of $1\times10^{-3}$ and a batch size of 64. Both the MLP and LSTM hidden dimensions are set to 256, and ELU activation is applied to all MLP layers. The temperature for the Gumbel-softmax sampler is fixed at 0.5. The prior variance of the trajectory $\sigma^{2}=5\times10^{-5}$.

\textbf{Results}. We compare the interaction graph predictions of our GVAE with those from other open-source state-of-the-art (SOTA) relational inference methods on the aforementioned swarm data. The results are shown in Table~\ref{table_acc_pr}. While all the methods predict both the trajectories and interaction graph edges, our focus is on graph prediction. We present a comparison of graph edge prediction performance in terms of precision, recall, and F1 score. 

We first compare our method to the linear correlation of the velocities between two agents (denoted as ``VC" in Table~\ref{table_acc_pr}), which is a straightforward, non-learning approach for determining inter-agent relations. Then we compare our method with other SOTA methods that predict interaction graph edges. GST\cite{liu2023intention} is based on mulit-head attention mechanism and is used in human interaction prediction for robot navigation. dNRI\cite{graber2020dynamic}, NRI\cite{kipf2018neural} and IMMA\cite{sun2022interaction} use a VAE to predict the agent trajectory and interaction graph edges. GAT\cite{velikovi2018graph} is a graph attention network and uses an attention mechanism to model the mutual relation.

\begin{figure}[thpb]
      \centering
      \includegraphics[scale=0.088]{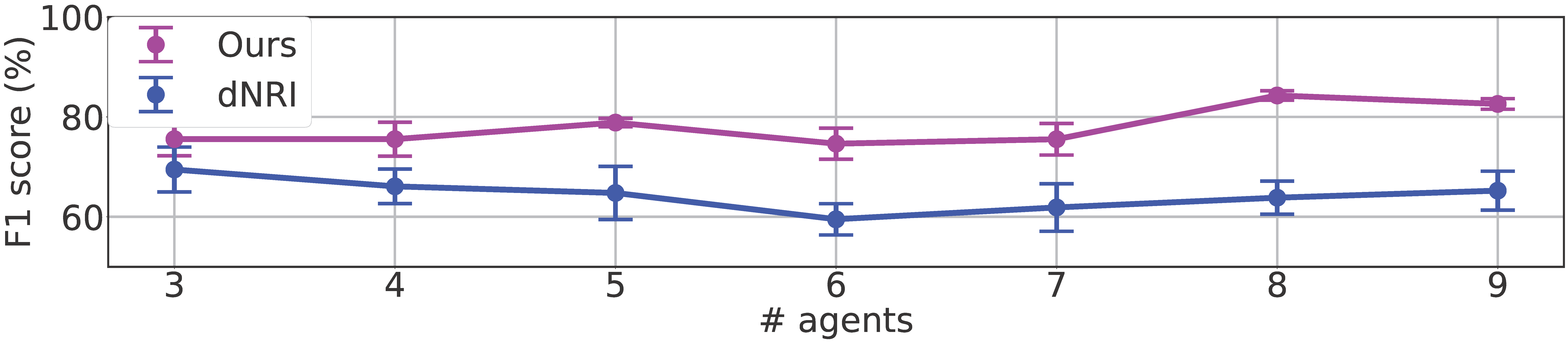}
      \caption{F1 score of graph prediction of our GVAE model compared with dNRI\cite{graber2020dynamic}.}
      \label{number}
   \end{figure}

Results in Table~\ref{table_acc_pr} show that our model gives better performance compared with other methods. Since scalability is an important aspect of swarming, we also compare our model's performance in terms of agent number. Fig.~\ref{number} shows the results. Our model consistently outperforms the strongest baseline (dNRI) across all agent numbers. The results demonstrate that our GVAE model shows improved accuracy for interaction graph edge prediction in the swarm setting, which lays the foundation for collective behavioral cloning in real-world swarm data.

\begin{table}[h]
\setlength\tabcolsep{2.2pt}
\caption{Comparison with SOTA methods of relational inference in terms of interaction graph edge prediction results. ``VC" stands for the linear velocity correlation. }
\begin{center}
\begin{tabular}{l c c c c}
\toprule
\textbf{Methods}   &   \textbf{Model Type}    &  \textbf{Precision (\%)}  & \textbf{Recall (\%)} & \textbf{F1 Score (\%)} \\
\midrule
VC                                  & Non-learning   & 67.89 ± 2.89     &  58.38 ± 2.21  & 62.72 ± 1.67        \\ 
GST\cite{liu2023intention}          & Transformer   &  58.85 ± 28.23   &  16.21 ± 6.33 &  25.18 ± 10.32   \\
dNRI\cite{graber2020dynamic}        & VAE   & 84.60 ± 2.96     & 53.46 ± 0.97          &  65.49 ± 1.12    \\
NRI\cite{kipf2018neural}            & VAE  & 61.25 ± 2.90     & 38.96 ± 0.66  & 47.59 ± 0.88               \\
GAT\cite{velikovi2018graph}         & Graph Net   &  49.89 ± 4.12    &  62.23 ± 3.28    & 55.38 ± 2.81        \\
IMMA\cite{sun2022interaction}       & VAE   & 26.19 ± 2.43     & 32.73 ± 2.72     & 29.09 ± 2.12        \\

\textbf{Ours }& VAE & \textbf{91.20  ± 2.48} & \textbf{73.84 ± 2.64} & \textbf{78.04 ± 2.80}  \\
\bottomrule
\end{tabular}
\end{center}
\label{table_acc_pr}
\end{table}

\begin{figure}[thpb]
   \centering
   \includegraphics[scale=0.264]{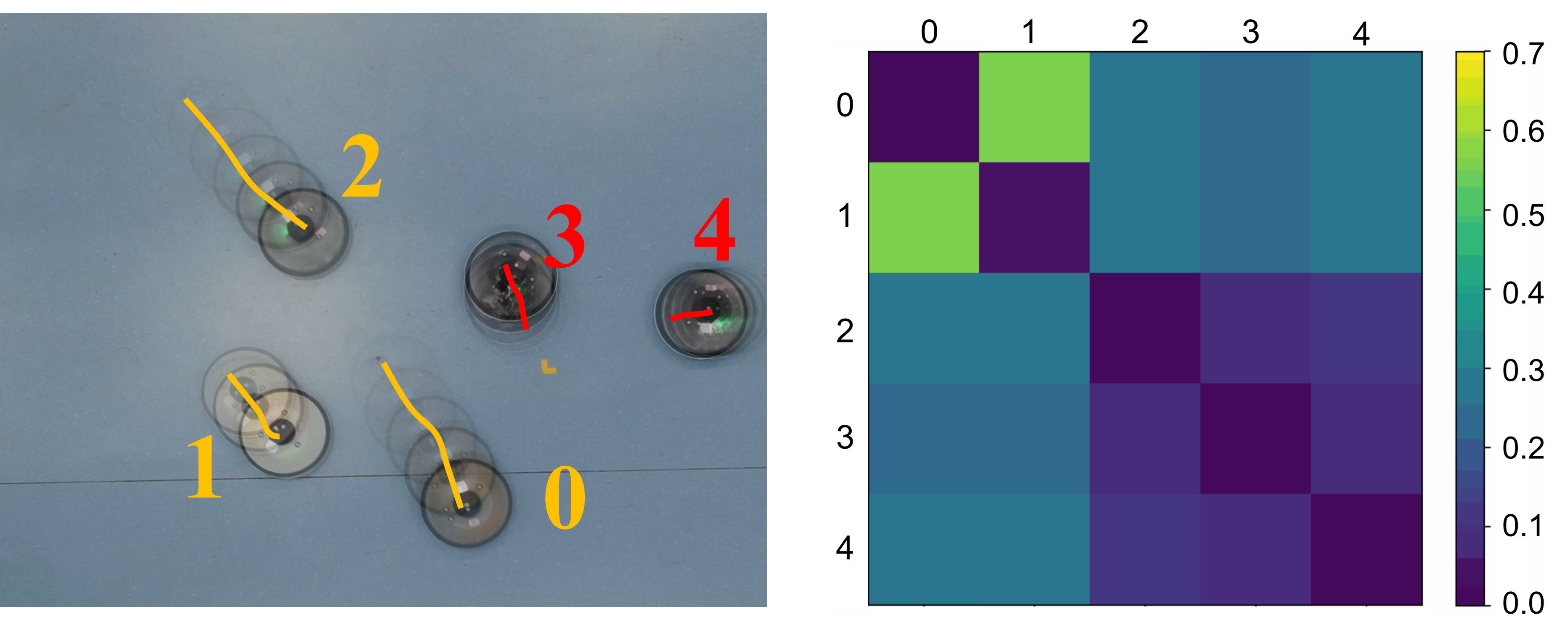}
   \caption{Robotic swarm trajectories and attention weights matrix from the GVAE. In the CBC experiment, Robot No.0 to No.2 belong to the same group and share larger attention weights, while No.3 and No.4 walk randomly and get lower attention weights. }
   \label{figure_pursue}
\end{figure}

   \begin{figure}[thpb]
      \centering
      \includegraphics[scale=0.151]{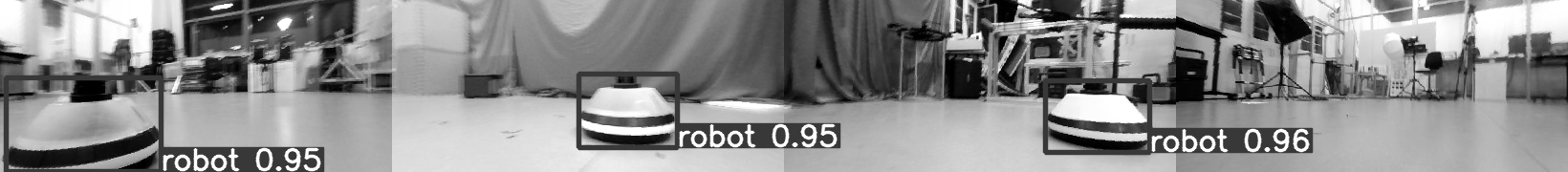}
      \includegraphics[scale=0.151]{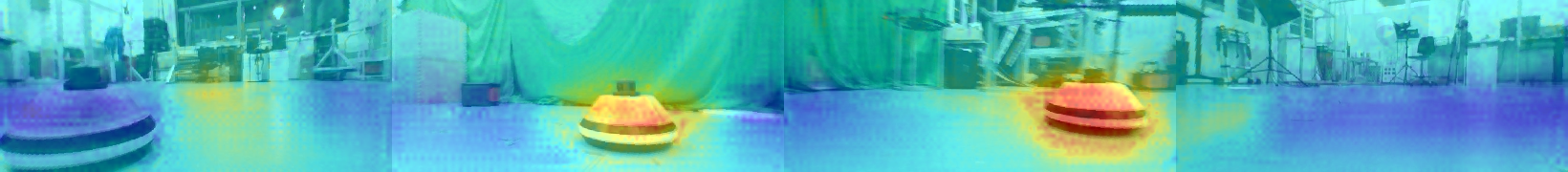}     
      \caption{Omnidirectional detection results of the visual-based robotic swarm. The first row shows the object detection results, and the second row shows the visual attention heat map. 
      }
      \label{figure_pursue}
 \end{figure}
\subsection{Results of Collective Behavioral Cloning}
\textbf{Overview}.
In this section, we deploy the algorithms on a real-world vision-based robot swarm to evaluate the performance of CBC. We use the control policy in Eq.~\ref{flocking} to generate swarm trajectory data. The action generated by $\pi$ is a 3-dimensional vector, representing the linear velocity $[v_{x},v_{y}]$ and angular velocity $\omega$. We use a total of 5 robots, divided into two groups. Group I consists of 3 robots, one robot moves in a predefined trajectory as the leader, and the other two robots are the followers. Group II consists of 2 robots performing a random walk. Robots within Group I are fully connected and continuously interact with each other, while there is no interaction between robots in Group I and Group II.
We use a two-layer MLP with 512 hidden units as $\hat{\pi}$ to clone the expert policy $\pi$. The input to the MLP consists of the estimated relative 2D velocities and positions of the interacting robots from the PLKF. The input dimension is set to $4 \times N$, where $N$ is the maximum number of robots in the experiments. When the number of interacting robots is less than $N$, the remaining part of the input tensor is padded with the states of the last interacting robot.
After generating the flocking trajectory data, we use GVAE to learn the interaction graph edge. With the learned local interaction graph edge, we use the projection approach introduced in Section III-C to generate heat map labels. Then with the heat map labels, the visual attention network is trained for interaction agent selection within the image. 

\textbf{System Implementation}
We test our framework on a real-world robotic swarm system. Each robot is equipped with a set of 4 cameras, oriented in different directions to provide an omnidirectional view. The onboard computer is an Nvidia Jetson Xavier NX (6 ARM cores, 8GB memory, and 21 TOPS of AI computing power). Ground truth trajectories of the robots are provided by a motion tracking system. To accelerate onboard inference and manage models efficiently, we utilize Nvidia TensorRT for inference acceleration and Triton for multi-model inference scheduling.

\begin{table}[h]
\setlength\tabcolsep{2.5pt}
\caption{Comparison of metrics for collective behavioral cloning with different interaction robot selection methods. Ours uses the learned interaction graph edge and visual attention model, which shows the best performance.}
\begin{center}
\begin{tabular}{l c c c  }
\toprule
\textbf{Interaction Methods}    &  \textbf{AE}     &  \textbf{TE} [m] &\textbf{AMD} [m]  \\
\midrule
Vision-Only             &     0.092 ± 0.033        &            3.01 ± 0.89                &  1.65 ± 0.49                  \\
Range-Based             &     0.051 ± 0.020       &             1.88 ± 0.55                &  0.95 ± 0.40                \\
K-Nearest               &     0.060 ± 0.021               &             1.79 ± 0.68                &  2.01 ± 0.30            \\
\textbf{Ours (Learning-based)}   &     \textbf{0.028 ± 0.016}       &            \textbf{ 0.95 ± 0.40 }               & \textbf{ 0.85 ± 0.31}       \\  

\bottomrule
\end{tabular}
\end{center}
\label{tableAE}
\end{table}

\textbf{Results of CBC}. The goal of the CBC experiment is to train a policy $\hat{\pi}$ to replicate the actions of $\pi_{i}$. We evaluate the performance using three metrics: action error (AE), trajectory error (TE), and average mutual distance (AMD). AE measures the mean error between the expert policy $\pi$ and the cloned policy $\hat{\pi}$; TE represents the accumulated trajectory error between the behavior generated by $\hat{\pi}$ and $\pi$; and AMD calculates the average mutual distance among robots in the same group during the experiment. The results are shown in Table~\ref{tableAE}, which demonstrates that our method exhibits the smallest action error, trajectory error, and average mutual distance. This is because our approach leverages the learned local interaction graph and a visual attention model to select interacting robots, whereas other methods rely on hand-crafted interaction rules, which are not consistent with the real interaction graph observed in the expert-generated swarm data.

\begin{figure}[thpb]
      \centering
      \includegraphics[scale=0.135]{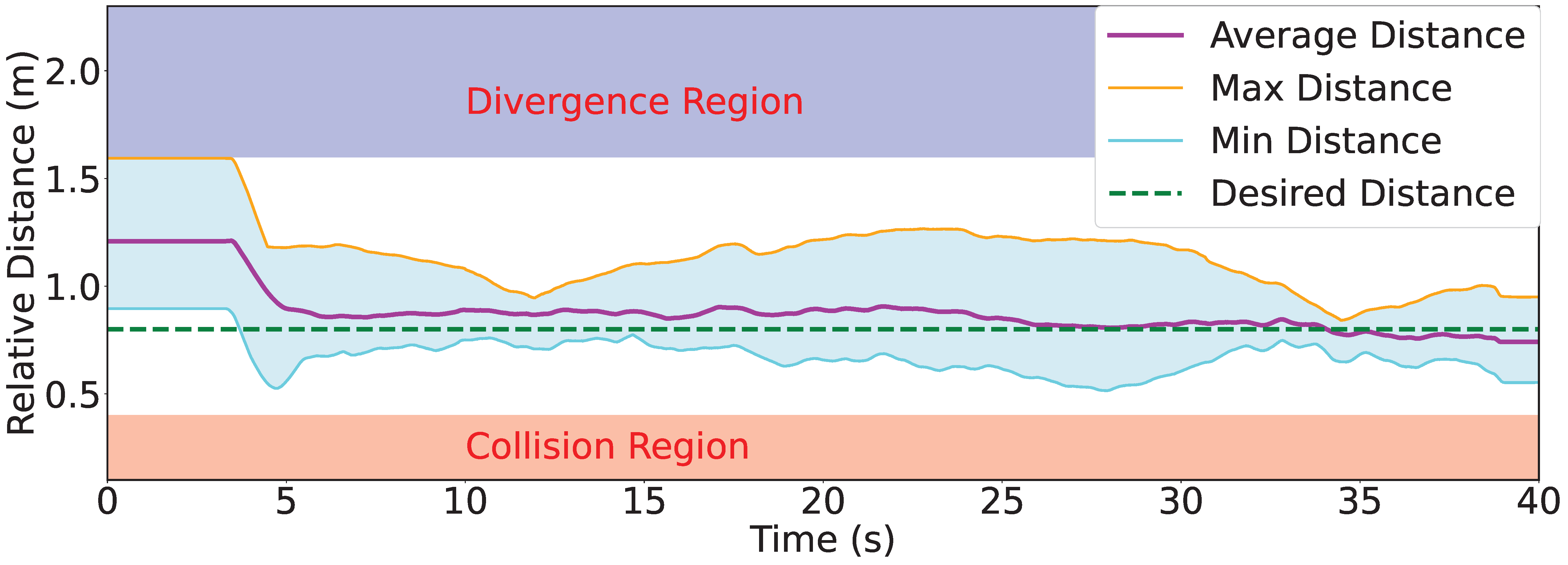}
      \caption{Relative distance during a single circular motion experiment. Collision and divergence thresholds are also indicated. The learned control policy and interaction graph keep all robots maintaining a desired mutual distance. }
      \label{relative_distance_curve}
   \end{figure}

\begin{figure}[thpb]
      \centering
      \includegraphics[scale=0.117]{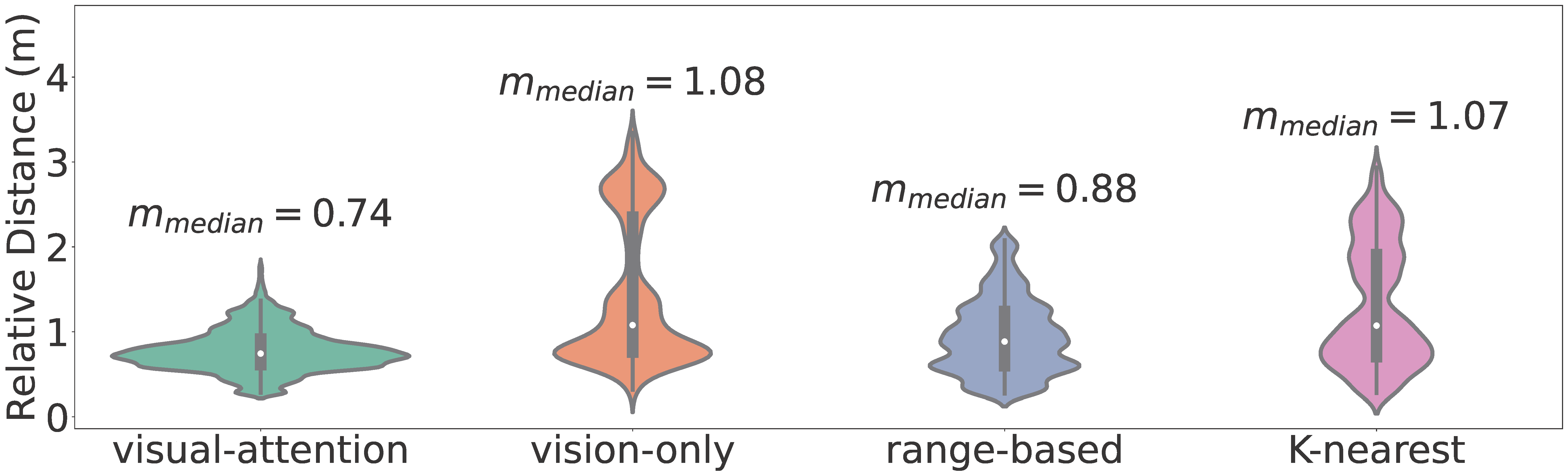}
      \caption{Comparison of relative distances among agents. Our approach with the learned interaction graph results in more stable and coherent behavior, maintaining relative distances around the desired value. In contrast, other target selection methods often exhibit irregular behavior with larger relative distances. }
      \label{bar}
   \end{figure}

Fig.~\ref{relative_distance_curve} depicts the average relative distance among agents of Group I during one experiment episode. The
desired relative distance (0.8 m) is defined by the expert policy $\pi$.
The curve indicates that the policy learned by our CBC framework can keep the swarm coherent while avoiding collision. 
Fig.~\ref{bar} shows the results of relative distance distribution during the experiment. While other methods often
lead the swarm to diverge, our method with the learned interaction selection mechanism keeps the swarm coherent at the desired mutual distance. This is because the visual-attention model actively filters task-specific robots to interact with, whereas other methods may include irrelevant ones, as all robots appear identical. The robotic flocking experiment involving two groups demonstrates that our CBC framework achieves more robust and stable performance, owing to the learned interaction graph.
Table~\ref{tableRunTime} lists the onboard running statistics of primary operating modules. The statistics demonstrate that our visual-attention-based swarm framework can be effectively deployed on real-world robotic systems with limited computational resources.

\textbf{Experiment Details}. In the experiment, the YOLOv5 and U-Net are converted to TensorRT model with half-precision (FP16) mode. YOLOv5 is trained on 1,996 simulation images and 3,281 real-world images for 100 epochs, achieving a mAP50 of 92\%. The visual attention model is a five-level U-Net\cite{ronneberger2015u} with (32,64,128,256,512) convolutional filters of kernel size (3,3,3,5,5). The scaling factor $\lambda$ for heatmap generation is set to 20. The final control command operates at a frequency of 30 Hz. 
The maximum interaction range of the range-based method is 1.6 m for comparison. Since we divide the robots into 2 groups and the maximum number of robots in each group is 3, the number of selected targets of the K-nearest is set at 2. The GVAE achieves a F1 score of 78\%.
The PLKF achieves an estimation accuracy of 0.078 ± 0.035 m within a detection range of 3 m. The variance of bearing and angle observations in the PLKF is determined based on the ground truth of the bounding box. The process noise of the PLKF is set as follows: $\sigma_{v}^2=0.1$, $\sigma_{p}^2=0.05$. The visual measurement noise of the PLKF is set as follows: $\sigma_{\boldsymbol{\mu}}^2=1\times10^{-6}$, $\sigma_{\omega}^2=1\times10^{-4}$. 

\begin{table}[h]
\setlength\tabcolsep{3pt}
\caption{Onboard running statistics. 100\% CPU usage refers to a single ARM core of the Jetson Xavier NX.}
\begin{center}
\begin{tabular}{l c c  c}
\toprule
\textbf{Modules}    &  \textbf{Input Shape}      &   \textbf{Running time (ms)}  &  \textbf{CPU (\%)}\\
\midrule
Image Concatenation &    $4\times320\times640$    &  4.28  ± 0.27  & 15.0         \\
Object Detection      &    $1\times64\times320$     &  10.90 ± 5.30   & 24.0         \\
Visual Attention    &    $1\times45\times410$     &  14.42 ± 6.41   & 54.6         \\
PLKF                &         N/A                 &  0.93 ± 0.11    & 8.9          \\
Robot Controller    &         N/A                 &  0.59 ± 0.09    & 3.9           \\
Triton Server       &         N/A                 &    N/A          & 25.5          \\

\bottomrule
\end{tabular}
\end{center}
\label{tableRunTime}
\end{table}

\section{Conclusions}
In this paper, we introduce a collective behavioral cloning (CBC) framework to learn the interaction graph and control policies of a swarm system, given its trajectory data. To capture the underlying interaction graph, we propose a Graph Variational Autoencoder (GVAE) model. Compared to previous models, our GVAE shows improved accuracy in predicting the interaction graph edges. We employ imitation learning to learn the control policy. The CBC framework is then deployed in a real-world, vision-based robot swarm system, where a visual attention network is used as a proxy for interaction neighbor selection. Our experiments demonstrate that the CBC framework can effectively learn both the interaction graph and control policy, leading to smaller action and trajectory errors. Future work may involve using this framework to study complex behaviors in humans or natural swarms.

\bibliographystyle{ieeetr}
\bibliography{root}
\end{document}